\documentclass[sigconf]{acmart}
\pdfoutput=1
\usepackage{booktabs} 
\usepackage{algorithm}
\usepackage{tabularx}

\setcopyright{rightsretained}
\copyrightyear{2020}
\acmYear{2020}
\setcopyright{rightsretained}
\acmConference[GECCO '20]{Genetic and Evolutionary Computation Conference}{July 8--12, 2020}{Canc\'{u}n, Mexico}
\acmBooktitle{Genetic and Evolutionary Computation Conference (GECCO '20), July 8--12, 2020, Canc\'{u}n, Mexico}
\acmDOI{10.1145/3377930.3390157}
\acmISBN{978-1-4503-7128-5/20/07}

\begin{document}
\title{Genetic programming approaches to learning fair classifiers}


\author{William La Cava}
\authornote{Corresponding Author}
\orcid{1234-5678-9012}
\affiliation{%
  \institution{University of Pennsylvania}
  \streetaddress{3700 Hamilton Walk}
  \city{Philadelphia} 
  \state{PA} 
  \postcode{19104}
}
\email{lacava@upenn.edu}

\author{Jason H. Moore}
\affiliation{%
  \institution{University of Pennsylvania}
  \streetaddress{3700 Hamilton Walk}
  \city{Philadelphia} 
  \state{PA} 
  \postcode{19104}
}
\email{jhmoore@upenn.edu}

\renewcommand{\shortauthors}{La Cava \& Moore}

\begin{abstract}
    Society has come to rely on algorithms like classifiers for important decision making, giving rise to the need for ethical guarantees such as fairness.
    Fairness is typically defined by asking that some statistic of a classifier be approximately equal over protected groups within a population.
    In this paper, current approaches to fairness are discussed and used to motivate algorithmic proposals that incorporate fairness into genetic programming for classification.
    We propose two ideas.
    The first is to incorporate a fairness objective into multi-objective optimization.
    The second is to adapt lexicase selection to define cases dynamically over intersections of protected groups.
    We describe why lexicase selection is well suited to pressure models to perform well across the potentially infinitely many subgroups over which fairness is desired.
    We use a recent genetic programming approach to construct models on four datasets for which fairness constraints are necessary, and empirically compare performance to prior methods utilizing game-theoretic solutions.
    Methods are assessed based on their ability to generate trade-offs of subgroup fairness and accuracy that are Pareto optimal.
    The result show that genetic programming methods in general, and random search in particular, are well suited to this task.
\end{abstract}

%
%



\maketitle

\section{Introduction}
Machine learning (ML) models that are deployed in the real world can have serious effects on peoples' lives.
In impactful domains such as lending~\cite{hardt_equality_2016}, college admissions~\cite{marcinkowski_implications_2020}, criminal sentencing~\cite{corbett-davies_algorithmic_2017,berk_fairness_2018}, and healthcare~\cite{gianfrancesco_potential_2018,zink_fair_2019}, there is increasing concern that models will behave in unethical ways~\cite{kearns_ethical_2019}.  
This concern has led ML researchers to propose different measures of fairness for constraining and/or auditing classification models~\cite{dwork_fairness_2012}.
However, in many cases, desired notions of fairness require exponentially many constraints to be satisfied, making the problems of learning fair models, and also checking for fairness, computationally hard~\cite{kearns_preventing_2017}.  
For this reason search heuristics like genetic programming (GP) may be useful for finding approximate solutions to these problems. 

This paper is, to our knowledge, the first foray into incorporating fairness constraints into GP.
We propose and study two methods for learning fair classifiers via GP-based symbolic classification.
Our first proposal is a straightforward one: to add a fairness metric as an objective to multi-objective optimization~\cite{schoenauer_fast_2000}.
This fairness metric works by defining \textit{protected groups} within the data, which match individuals having a specific value of one protected attribute, e.g. ``female" for a sex attribute.  
Unfortunately, simple metrics of fairness do not capture fairness over rich subgroups and/or intersections of groups - that is, over multiple protected attributes that intersect in myriad ways. 
With this in mind, we propose an adaptation of lexicase selection~\cite{la_cava_probabilistic_2018} designed to operate over randomized sequences of fairness constraints. 
This algorithm draws a connection between these numerous fairness constraints and the way in which lexicase samples fitness cases in random sequences for parent selection. 
We illustrate the ability of lexicase to sample the space of group intersections in order to pressure models to perform well on the intersections of groups that are most difficult in the current population. 
In our experiments, we compare several randomized search heuristics to a recent game-theoretic approach to capturing subgroup fairness.
The results suggest that GP methods can produce Pareto-efficient trade-offs between fairness and accuracy, and that random search is a strong benchmark for doing so.

In the following section, we describe how fairness has been approached in the ML community and the challenges that motivate our study.
Section~\ref{s:methods} describes the algorithms we propose in detail, and Section~\ref{s:exp} describes the experiment we conduct on four real-world datasets for which fairness concerns are pertinent. 
We present resulting measures of performance, statistical comparisons, and example fairness-accuracy trade-offs in Section~\ref{s:results}, followed finally by a discussion of what these results entail for future studies.

\section{Background}
\label{s:background}

Incorporating notions of fairness into ML is a fairly new idea~\cite{pedreshi_discrimination-aware_2008}, and early work in the field is reviewed in Chouldechova and Roth~\cite{chouldechova_frontiers_2018}.
Algorithmic unfairness may arise from disparate causes, but often has to do with the properties of the data used to train a model. 
One major cause of bias is that data are often collected from unequal demographics of a population. 
In such a scenario, algorithms that minimize average error over all samples will skew towards fitting the majority population, since this leads to lower average error. 
One way to address this problem is to train separate models for separate demographic populations. 
In some scenarios, this method can reduce bias, but there are two main caveats, expounded upon in~\cite{thomas_preventing_2019}.
First, some application areas explicitly forbid demographic data to be used in prediction, meaning these models could not be deployed. 
The second, and more general, concern is that we may want to protect several sensitive features of a population (e.g., race, ethnicity, sex, income, medical history, etc.).
In those cases, dividing data beforehand is non-trivial, and can severely limit the sample size used to train each model, leading to poor performance. 

There is not a single agreed-upon definition of fairness for classification. 
The definitions put forth can be grouped into two kinds: {\it statistical} fairness, in which we ask a classifier to behave approximately equally on average across protected groups according to some metric; and {\it individual} fairness, in which we ask a classifier to perform similarly on similar pairs of individuals~\cite{dwork_fairness_2012}. 
For this paper, we focus on statistical fairness, especially equality of false positive (FP), false negative (FN), and accuracy rates among groups.  
We essentially ask that the classifier's errors be distributed among different protected groups as evenly as possible. 

Fairness constraints have been proposed for classification algorithms, for example by regularization~\cite{dwork_fairness_2012,berk_convex_2017}, model calibration~\cite{hardt_equality_2016}, cost-sensitive classification~\cite{agarwal_reductions_2018}, and evolutionary multi-objective optimization~\cite{quadrianto_recycling_2017}. 
For the most part, literature has focused on providing guarantees over a small number of protected groups that represent single attributes - for example, race and sex. 
However, a model that appears fair with respect to several individual groups may actually discriminate over specific {\it intersections} or conjunctions of those groups. 
Kearns et al.~\cite{kearns_preventing_2017} refers to this issue as ``fairness gerrymandering". 
To paraphrase Example 1.1 of their work~\cite{kearns_preventing_2017}, imagine a classifier that exhibits equivalent error rates according to two protected groups: a race feature taking values in \{``black", ``white"\} and, separately, a sex feature taking values in \{``male", ``female"\}. 
This seemingly fair classifier could actually be producing 100\% of its errors on black males and white females. 
In such a case the classifier would appear fair according to the individual race and sex groups, but unfair with respect to their conjunction. 

If we instead wish to learn a classifier that is fair with respect to both individual groups defined over single attributes and boolean conjunctions of those groups, a combinatorial problem arises.
For $p$ protected attributes, we have to both learn and check for fairness over $2^p$ groups. 
It turns out that the problems of auditing a classifier for fairness over boolean conjunctions of groups (as well as other group definitions) is computationally hard in the worst case, as is the classification problem~\cite{kearns_preventing_2017}.

Kearns et al.~\cite{kearns_preventing_2017} proposed a heuristic solution to the problem of learning a classifier with rich subgroup fairness constraints by formulating it as a two-player game in which one player learns a classifier and the other learns to audit that classifier for fairness. 
They empirically illustrated the trade-off between fairness violations and model accuracy on four real-world problems~\cite{kearns_empirical_2018}. 
In our study, we build upon their work by using their fairness auditor to compare performance of models on the same datasets. 
In their study, Kearns et al. focused on algorithmic characterization by reporting fairness and accuracy on the training samples.
Conversely, we are interested in the generalization performance of the learned classification models; therefore we conduct our comparisons over cross-validated predictions, rather than reporting in-sample. 

Our interest in applying GP to the problem of fair classification is motivated by three observations from this prior work. 
First, given that the learning and auditing problems for rich subgroup fairness are hard in the worst case means that a heuristic method such as GP may be able to provide approximate solutions with high utility, and therefore it is worth an empirical analysis. 
Second, many authors note the inherent trade-off that exists between fairness and accuracy~\cite{hardt_equality_2016,kearns_empirical_2018} and the need for Pareto-efficient solution sets. 
Multi-objective optimization methods that are typically used in GP (e.g., NSGA2~\cite{schoenauer_fast_2000}) are well-suited to handle competing objectives during search.
Finally, we note that demographic imbalance, one of the causes of model unfairness, is a problem due to the use of average error for guiding optimization. 
However, recent semantic selection methods~\cite{liskowski_comparison_2015} such as $\epsilon$-lexicase selection~\cite{la_cava_epsilon-lexicase_2016} are designed specifically to move away from scalar fitness values that average error over the entire training set. 
The original motivation behind these GP methods is to prevent the loss of candidate models in the search space that perform well over difficult subsets of the data~\cite{la_cava_epsilon-lexicase_2016}. 
Furthermore, we hypothesize that $\epsilon$-lexicase selection may be adapted to preserve models that perform well over structured subgroups of the protected attributes as well.  

\section{Methods}
\label{s:methods}
We start with a dataset of triples, $\mathcal{D} = \{(\mathbf{x}_i,\mathbf{x}_i',y_i)\}_{i = 1}^{m}$, containing $m$ examples. 
Our labels $y \in \{0,1\}$ are binary classification assignments and $\mathbf{x}$ is a vector of $d$ features.  
In addition to $\mathbf{x}$, we have a vector of $p$ sensitive features, $\mathbf{x}'$, that we wish to protect via some fairness constraint. 
It is worth mentioning that for the purposes of this study, $\mathbf{x}$ contains $\mathbf{x}'$, meaning that the learned classifier has access to the sensitive attribute observations in prediction; this is not always the case (e.g. ~\cite{thomas_preventing_2019}). 

We also define protected groups $\mathcal{G}$, where each $g \in \mathcal{G}$ is an indicator function\footnote{We use $\mathbf{1}\{\}$ to denote indicator functions.}, mapping a set of sensitive features $\mathbf{x}'$ to a group membership. 
It is useful to define a simple set of protected groups that correspond to the unique levels of each feature in $\mathbf{x}'$. 
We will call the set of these simple groups $\mathcal{G}_0$. 
As an example, imagine we have two sensitive features corresponding to race and sex: $x_1' \in \{\text{black, white}\}$ and $x_2' \in \{\text{male, female}\}$. 
Then $\mathcal{G}_0$ would consist of four groups:
\begin{align*}
    \mathcal{G}_0 = &\{ g_1(\mathbf{x}') = \mathbf{1}\{x_1' = \text{black}\}, \\ 
                      &g_2(\mathbf{x}') = \mathbf{1}\{x_1' = \text{white}\}, \\ 
                      &g_3(\mathbf{x}') = \mathbf{1}\{x_2' = \text{male}\}, \\ 
                      &g_4(\mathbf{x}') = \mathbf{1}\{x_2' = \text{female}\} \}
\end{align*}
We make use of $\mathcal{G}_0$ in defining marginal fairness and in Algorithm~\ref{alg}.

We use a recent GP technique called FEAT~\cite{la_cava_learning_2019,la_cava_semantic_2019} that evolves feature sets for a linear model, in this case a logistic regression model. 
More details of this method are given in Section~\ref{s:exp}.
As in other GP methods, FEAT trains a population of individuals, $n \in \mathcal{N}$, each of which produces binary classifications of the form $n(\mathbf{x}) \in \{0,1\}$. 
The fitness of $n$ is its average loss over the training samples, denoted $f(n)$. 
We refer to the fitness of $n$ over a specific group of training samples as $f(n,g)$. 
With these definitions in mind, we can define the fairness of a classifier with respect to a particular group and fitness measure as:

\begin{equation}
    f\text{-Fairness}(n,g) = |f(n) - f(n,g)|
    \label{eq:fairness}
\end{equation}

FEAT uses logistic loss as its fitness during training, in keeping with its logistic regression pairing. 
However, we compare fairness on fitted models relative to the FP and FN rate, as in previous work~\cite{kearns_empirical_2018,agarwal_reductions_2018}.

\subsection{Multi-objective Approach}

A straightforward way to incorporate fairness into FEAT is to add it as an objective to a multi-objective optimization algorithm like NSGA2.
We use the term {\it marginal fairness} to refer to the first-level fairness of a model defined over simple groups $\mathcal{G}_0$:
\begin{equation}
    f\text{-Marginal Fairness}(n,\mathcal{G}_0) = \frac{1}{|\mathcal{G}_0|}\sum_{g \in \mathcal{G}_0}{f\text{-Fairness(n,g)}}
    \label{eq:marg_fair}
\end{equation}

A challenge with using fairness as an objective is the presence of a trivial solution: a model that produces all 1 or all 0 classifications has perfect fairness, and will easily remain in the population unless explicitly removed. 

A major shortcoming of optimizing Eqn.~\ref{eq:marg_fair} is that it does not pressure classifiers to perform well over group intersections, and is therefore susceptible to fairness gerrymandering, as described in Section~\ref{s:background}.
Unfortunately, it is not feasible to explicitly audit each classifier in the population each generation over all possible combinations of structured subgroups. 
While an approximate, polynomial time solution has been proposed~\cite{kearns_preventing_2017,kearns_empirical_2018}, we consider it too expensive to compute in practice each iteration on the entire set of models.  
For these reasons, we propose an adaptation of lexicase selection~\cite{spector_assessment_2012} to handle this task in the following section.

\subsection{Fair Lexicase Selection}

Lexicase selection is a parent selection algorithm originally proposed for program synthesis tasks~\cite{helmuth_solving_2014} and later regression~\cite{la_cava_epsilon-lexicase_2016}. 
Each parent selection event, lexicase selections filters the population through a newly randomized ordering of ``cases", which are typically training samples. 
An individual may only pass through one of these cases if it has the best fitness in the current pool of individuals, or alternately if it is within $\epsilon$ of the best for $\epsilon$-lexicase selection. 
The filtering process stops when one individual is left (and is selected), or when it runs out of cases, resulting in random selection from the remaining pool. 

Although different methods for defining $\epsilon$ have been proposed, we use the most common one, which defines $\epsilon$ as the median absolute deviation ($\lambda$) of the loss ($\ell$) in the current selection pool:
\footnote{Defining $\lambda$ relative to the current selection pool is called ``dynamic $\epsilon$-lexicase selection" in~\cite{la_cava_probabilistic_2018}.}

\[ \lambda(\ell(n), n \in \mathcal{S}) = \text{median}(|\ell(n) - \text{median}(\ell(n))|), n \in \mathcal{S} \]

Lexicase selection has a few properties worth noting that are discussed in depth in~\cite{la_cava_probabilistic_2018}. 
First, it takes into account case ``hardness", meaning training samples that are very easy to solve apply very little selective pressure to the population, and vice versa. 
Second, lexicase selection selects individuals on the Pareto front spanned by the cases; this means that, in general, it is able to preserve individuals that only perform well on a small number of hard cases (i.e. specialists~\cite{helmuth_lexicase_2019}).
Third, and perhaps most relevant to rich subgroup fairness, lexicase selection does not require each individual to be run on each case/sample, since selection often chooses a parent before the cases have been exhausted~\cite{la_cava_epsilon-lexicase_2016}. 
The worst case complexity of parent selection is $O(|\mathcal{N}|^2m)$, which only occurs in a semantically homogeneous population.

Because of the third point above, we can ask for lexicase selection to audit classifiers over conjunctions of groups without explicitly constructing those groups beforehand. 
Instead, in {\it fair lexicase} (FLEX, detailed in Alg.~\ref{alg}), we define ``cases" to be drawn from the simple groups in $\mathcal{G}_0$. 
A randomized ordering of these groups, i.e. cases, thereby assesses classifier performance over a conjunction of protected attributes. 
By defining cases in this way, selective pressure moves dynamically towards subgroups that are difficult to solve.
For any given parent selection event, lexicase only needs to sample as many groups as are necessary to winnow the pool to one candidate, which is at most $|\mathcal{G}_0|$.
Nonetheless, due to the conditional nature of case orderings, and the variability in case depth and orderings, lexicase effectively samples ${|\mathcal{G}_0|!}$ combinations of protected groups.

\begin{figure}
    \includegraphics[width=\columnwidth]{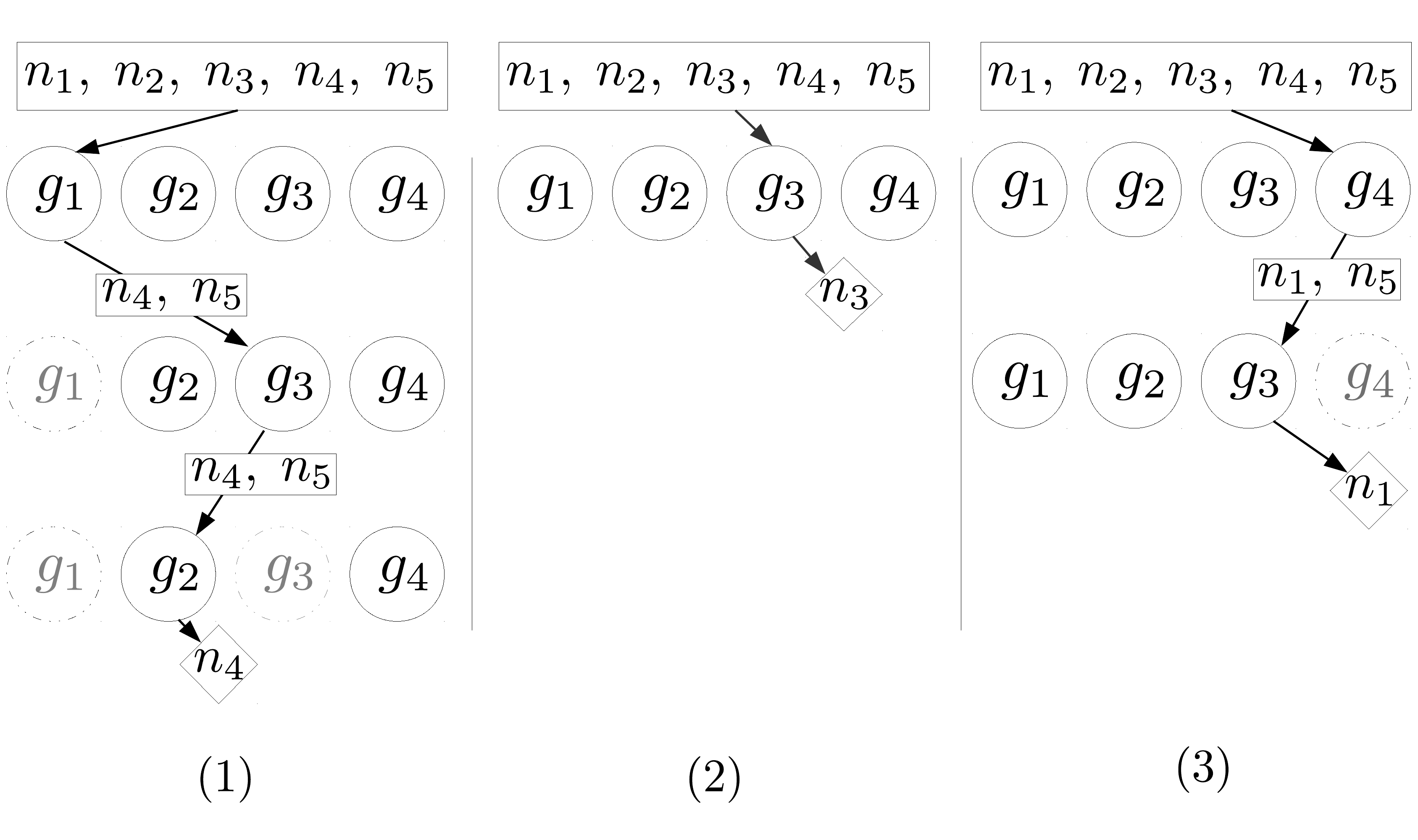}
    \caption{
        Three example selection events with FLEX, with a population $\mathcal{N} = \{n_1,\;\dots,\;n_5\}$ and protected groups $\mathcal{G} = \{g_1,\;\dots,\;g_4\}$. 
        Parent selection 1) selects on the conjunction of $g_1$, $g_3$, and $g_2$ to select $n_4$.
        Note that $g_3$ exerts no selection pressure because $n_4$ and $n_5$ both perform well on it. 
        2) Here a single group, $g_3$, is enough to winnow the population to $n_3$, which is selected.
        3) Selection on $g_4$ and $g_3$ to select $n_1$. 
        Gray cases are paths that have already been visited for a given selection event. 
}
\label{fig:example}
\end{figure}

An illustration of three example selection events is shown in Figure~\ref{fig:example}. 
These events illustrate that FLEX can select on different sequences of groups and different sequence lengths, while also taking into account the easiness or hardness of the group among the current selection pool. 

A downside of FLEX versus the multi-objective approach is that it is not as clear how to pressure for both fairness and accuracy among cases. 
On one hand, selecting for accuracy uniformly over many group definitions could lead to fairness, but it may also preserve markedly unfair, and therefore undesirable, models.
We address this issue by allowing both case definitions to appear with equal probability. 
This choice explains the random coin flip in Alg.~\ref{alg}.

\begin{algorithm}
    \caption{{\bf : Fair $\epsilon$-Lexicase Selection (FLEX)} 
    applied to individuals $n \in \mathcal{N}$ with loss $f(n,g)$ over protected groups $g \in \mathcal{G}_0$.}\label{alg}
\noindent{\small
\begin{tabularx}{\columnwidth}{lX} 
    \texttt{Selection}($\mathcal{N,G}_0$) 	:						&	\\
    \hspace{1em} $\mathcal{P} \leftarrow \emptyset$ & $\lozenge$ parents \\
\hspace{1em} \textbf{do} \text{N} \textbf{times}: &\\
\hspace{1em}  \hspace{1em} $\mathcal{P} \leftarrow \mathcal{P} \; \cup \; $\texttt{GetParent}($\mathcal{N,G}_0$) & 
$\lozenge$ add selection to $\mathcal{P}$ \\
\\
\texttt{GetParent}($\mathcal{N,G}_0$) 	:						&	\\
\hspace{1em}	$\mathcal{G}' \leftarrow \mathcal{G}_0$	& $\lozenge$	protected groups \\
\hspace{1em}	$S \leftarrow \mathcal{N}$	& $\lozenge$	 selection pool\\
\hspace{1em}	\textbf{while} $|\mathcal{G}'| >0$ \textbf{and} $|\mathcal{S}|>1$:						&	\\
\hspace{1em}\hspace{1em}	$g$ $\leftarrow$ random choice from $\mathcal{G'}$ & $\lozenge$ pick random group \\
\hspace{1em}\hspace{1em}	\textbf{if} random number $\in [0,1] < 0.5$ \textbf{then} \\
\hspace{1em}\hspace{1em}\hspace{1em}	$\ell(n)$ $\leftarrow$ $f(n,g)$ \textbf{for} $n \in \mathcal{S}$ & $\lozenge$ loss over group \\
\hspace{1em}\hspace{1em}	\textbf{else} \\
\hspace{1em}\hspace{1em}\hspace{1em}	$\ell(n)$ $\leftarrow$ $f$-Fairness$(n,g)$ \textbf{for} $n \in \mathcal{S}$ & $\lozenge$ group fairness \\
\hspace{1em}\hspace{1em}	$\ell^*$ $\leftarrow$ min $\ell(n)$ \textbf{for} $n \in \mathcal{S}$ 	&$\lozenge$ min fitness in pool\\
\hspace{1em}\hspace{1em}	$\epsilon \leftarrow \lambda(\ell(n),n \in \mathcal{S})$ 	&$\lozenge$ deviation of fitnesses \\
\hspace{1em}\hspace{1em}	\textbf{for} $n \in \mathcal{S}$:&\\
\hspace{1em}\hspace{1em}\hspace{1em}	 \textbf{if} $\ell(n)$ $> \ell^*+\epsilon$ \textbf{then}&\\
\hspace{1em}\hspace{1em}\hspace{1em}\hspace{1em} $\mathcal{S} \leftarrow \mathcal{S} \setminus \{n\}$	 &$\lozenge$ filter selection pool \\
\hspace{1em}\hspace{1em}	$\mathcal{G'} \leftarrow \mathcal{G'} \setminus \{g\}$				& $\lozenge$ remove $g$ \\
\hspace{1em} \textbf{return} random choice from $\mathcal{S}$
\end{tabularx}
}
\end{algorithm}

\section{Experiments}
\label{s:exp}
 
We conduct our experiment on four datasets used in previous related work~\cite{kearns_empirical_2018}. 
These datasets and their properties are detailed in Table~\ref{tbl:datasets}. 
Each of these classification problems contain sensitive information for which one would reasonably want to assure fairness.   
Two of the datasets concern models for admissions decisions (Lawschool and Student); 
The other two are of concern for lending and credit assessment: one predicts rates of community crime (Communities), and the other attempts to predict income level (Adult). 
For each of these datasets we used the same cleaning procedure as this previous work, making use of their repository (available here: \href{https://github.com/algowatchpenn/GerryFair}{github.com/algowatchpenn/GerryFair}).

\begin{table*}
    \centering
    \caption{Properties of the datasets used for comparison.}
    \label{tbl:datasets}
    \footnotesize
    \begin{tabularx}{\textwidth}{lllrrrXX}
        Dataset     &   Source (link)  &   Outcome &   Samples &   Features    & Sensitive features  & Protection Types   & Number of simple groups ($|\mathcal{G}_0|$) \\ \hline
        Communities &   \href{http://archive.ics.uci.edu/ml/datasets/communities+and+crime}{UCI}    
                    &   Crime rates     &   1994    &   122     &  18   &  race, ethnicity, nationality    &   1563    \\ 
        Adult       &   \href{https://archive.ics.uci.edu/ml/datasets/adult}{Census}    
                    &   Income          &   2020    &   98      &   7  &   age, race, sex    &   78    \\
        Lawschool   &   \href{https://eric.ed.gov/?id=ED469370}{ERIC}    
                    &   Bar passage     &   1823    &  17       &   4   &   race, income, age, gender   &   47 \\
        Student     &   \href{https://archive.ics.uci.edu/ml/datasets/student+performance}{Secondary Schools}    
                    &   Achievement     &   395     &   43      &   5  &   sex, age, relationship status, alcohol consumption  &   22  \\
        \hline
    \end{tabularx}    
\end{table*}

We compared eight different modeling approaches in our study, the parameters of which are shown in Table~\ref{tbl:algs}. Here we briefly describe the two main algorithms that are used. 

\paragraph{GerryFair}
First, we used the ``Fictitious Play" algorithm from~\cite{kearns_preventing_2017,kearns_empirical_2018}, trained for 100 iterations at 100 different levels of $\gamma$, which controls the trade-off between error and fairness. 
As mentioned earlier, GerryFair treats the problem of learning a fair classifier as a two player game in which one player, the classifier, is attempting to minimize error over weighted training samples, and the other player, the auditor, is attempting to find the subgroup within the classifier's predictions that produces largest fairness violation. 
The play continues for the maximum iterations or until the maximum fairness violation is less than $\gamma$. 
The final learned classifier is an ensemble of linear, cost-sensitive classification models. 
We make use of the auditor for validating the predictions of all compared models, so it is described in more detail in Section~\ref{s:auditor}.

\paragraph{FEAT}

\begin{figure}
    \includegraphics[width=0.8\columnwidth]{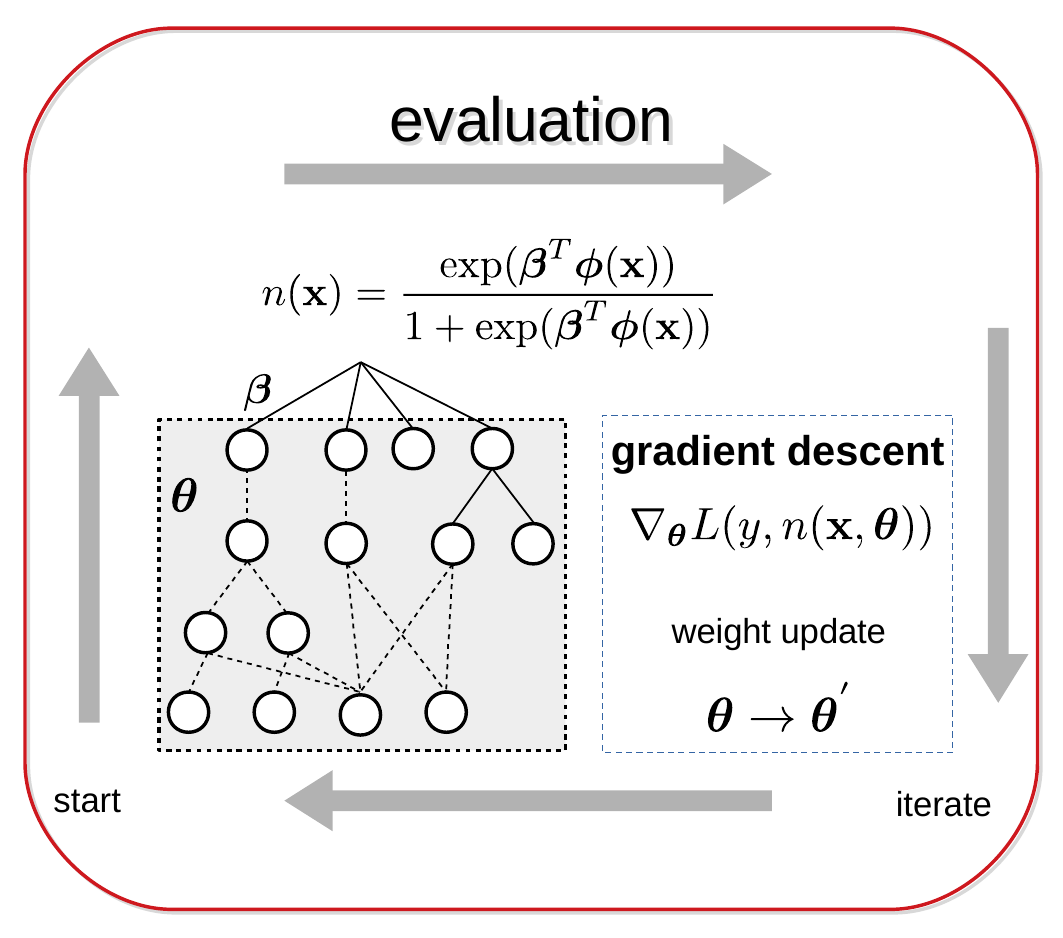}
    \caption{
        Diagram of the evaluation of a single FEAT individual, which produces a logistic regression model over program outputs $\phi$. 
        The internal weights $\boldsymbol{\theta}$ are trained via gradient descent each generation for a set number of iterations. 
}
\label{fig:feat}
\end{figure}

Our GP experiments are carried out using the Feature Engineering Automation Tool (FEAT), a GP method in which each individual model $n$ consists of a set of programs (i.e. engineered features) that are fed into a logistic regression model (see Figure~\ref{fig:feat}). 
This allows FEAT to learn a feature space for a logistic regression model, where the number of features is learned via the search process.
The features are comprised of continuous and boolean functions, including common neural network activation functions, as shown in Table 1 in~\cite{la_cava_learning_2019}.
We choose to use FEAT for this experiment because it performed well in comparison to other state-of-the-art GP methods on a battery of regression tests~\cite{la_cava_semantic_2019}.  
FEAT is also advantageous in this application to binary classification because it can be paired with logistic regression, which provides probabilistic outputs for classification. 
These probabilities are necessary for assessing model performance using certain measures such as the average precision score, as we will describe later in Eqn.~\ref{eq:aps}. 

FEAT trains models according to a common evolutionary strategy. 
This strategy begins with the construction of models, followed by selection for parents. 
The parents are used to produced offspring via mutation and crossover. 
Depending on the method used, parents and offspring may then compete in a survival step (as in NSGA2), or the offspring may replace the parents (LEX, FLEX). 
For further details of FEAT we refer the reader to~\cite{lacava_learning_2020} and to the github project (\href{http://github.com/lacava/feat}{github.com/lacava/feat}). 

We test six different selection/survival methods for FEAT, shown in Table~\ref{tbl:algs}. 
FLEX-NSGA2 is a hybrid of FLEX and NSGA2 in which selection for parents is conducted using FLEX and survival is conducted using the survival step of NSGA2. 
Each GP method was trained for 100 generations with a population of 100, except for Random, which returned the initial population.
These parameters were chosen to approximately match those of GerryFair, and to produce the same number of final models (100). 
However, since the GP methods are population-based, they train 100 models per generation (except Random).
GerryFair only trains two models per iteration (the classifier and the auditor); thus, at a first approximation we should expect the GP models aside from Random to require roughly 50 times more computation. 

In our experiments, we run 50 repeat trials of each method on each dataset, in which we split the data 50/50 into training and test sets. 
For each trial, we train models by each method, and then generate predictions on the test set over each returned model. 
Each trial is run on a single core in a heterogeneous cluster environment, consisting mostly of 2.6GHz processors with a maximum of 8 GB of RAM. 

There are inherent trade-offs between notions of fairness and accuracy that make it difficult to pick a definitive metric by which to compare models~\cite{kleinberg_inherent_2016}.
We compute several metrics of comparison, defined below.

\subsection{Auditing Subgroup Fairness}
\label{s:auditor}
In order to get a richer measure of subgroup fairness for evaluating classifiers, Kearns et al.~\cite{kearns_empirical_2018} developed an auditing method that we employ here for validating classifiers. 
The auditor uses cost-sensitive classification to estimate the group that most badly violates a fairness measure they propose, which we refer to as a subgroup FP- or FN- Violation. 
We can define this relative to FP rates as
\begin{eqnarray}
    \alpha_{FP}(g,\mathcal{P}) = Pr_{\mathcal{P}}[g(\mathbf{x'})=1, y=0] \nonumber \\
    \beta(n,g) = |FP(n) - FP(n,g)| \nonumber \\
    \text{FP-Violation}(n,g,\mathcal{P}) =  \alpha_{FP}(g) \beta(n,g,\mathcal{P})
    \label{eq:audit}
\end{eqnarray}
here, $\mathcal{P}$ is the distribution from which the data $\mathcal{D}$ is drawn.
In Eqn.~\ref{eq:audit}, $\alpha_{FP}(g, \mathcal{P})$ is estimated by the fraction of samples group $g$ covers, so that larger groups are more highly weighted. 
$\beta$ measures fairness equivalently to Eqn.~\ref{eq:fairness}. 
This metric can be defined equivalently for FN subgroup violations, and we report both measures in our experiments.
The auditing algorithm's objective is to return an estimate of the group $g$ with the highest FP- or FN-Violation, and this violation is used as a measure of classifier unfairness. 

\subsection{Measures of Accuracy}
In order to compare the accuracy of the classifiers, we used two measures. 
The first is accuracy, defined as
\begin{equation}
    Accuracy(n) = \frac{1}{m}\sum_{i}^m{\mathbf{1}[n(\mathbf{x}_i) = y_i]}
    \label{eq:accuracy}
\end{equation}

The second is average precision score\footnote{This is a pessimistic version of estimating area under the precision-recall curve. See \url{https://scikit-learn.org/stable/modules/generated/sklearn.metrics.average_precision_score.html}.}, 
which is the mean precision of the model at different classification thresholds, $t$.
APS is defined as 

\begin{equation} \label{eq:aps}
    APS(n) = \sum_t{(R_t(n) - R_{t-1}(n))P_t(n)}
\end{equation}
where $R(n) = Pr[n=1,y=1]/Pr[y=1]$ is the recall and $P(n) = Pr[n=1,y=1]/Pr[n=1]$ is the precision of $n(\mathbf{x})$. 

\subsection{Comparing Accuracy-Fairness Trade-offs}
It is well known that there is a fundamental trade-off between the different notions of fairness described here and classifier accuracy~\cite{hardt_equality_2016,berk_fairness_2018,kleinberg_inherent_2016}. 
For this reason, recent work has focused on comparing the Pareto front of solutions between methods~\cite{kearns_empirical_2018}. 
For GerryFair, this trade-off is controlled via the parameter $\gamma$ described in Table~\ref{tbl:algs}.
For the GP methods, we treat the final population as the solution set to be evaluated. 

In order to compare sets of solutions between methods, we compute the hypervolume of the Pareto front~\cite{fonseca_improved_2006} between competing pairs of accuracy objectives (Accuracy, APS) and fairness objectives (FP Subgroup Violation, FN Subgroup Violation).
This results in four hypervolume measures of comparison.
For two objectives, the hypervolume provides an estimate of the area of the objective space that is covered/dominated by a set of solutions. 
Thus, the hypervolume allows us to compare how well each method is able to characterize the fairness-accuracy trade-off~\cite{chand_evolutionary_2015}. 

\begin{table}
    \centering
    \caption{Settings for the methods in the experiments.}\label{tbl:algs}
    \begin{tabularx}{\columnwidth}{lX}
        Algorithm   &   Settings    \\ \toprule
        GerryFair~\cite{kearns_preventing_2017} &   iterations=100, $\gamma$= 100 values $\in [0.001,\;\dots,\;1]$, ml = logistic regression\\ 
		- GerryFairGB							& ``", ml = gradient boosting  \\\midrule
        FEAT~\cite{la_cava_learning_2019}        &  generations=100, pop size=100, max depth=6, max dim=20   \\
        - Tourn       &  selection: size 2 tournament selection     \\
        - LEX~\cite{la_cava_epsilon-lexicase_2016}         &  selection: $\epsilon$-lexicase selection    \\
        - FLEX (Alg.~\ref{alg})        &  selection: Fair $\epsilon$-lexicase selection    \\
        - NSGA2~\cite{schoenauer_fast_2000}       &  NSGA2 selection and survival \\
        - FLEX-NSGA2  &  selection: $\epsilon$-lexicase selection, survival: NSGA2 \\ 
        - Random  &  return initial random population \\ \bottomrule
    \end{tabularx} 
\end{table}

\section{Results}
\label{s:results}
\begin{figure}
    \includegraphics[width=\columnwidth]{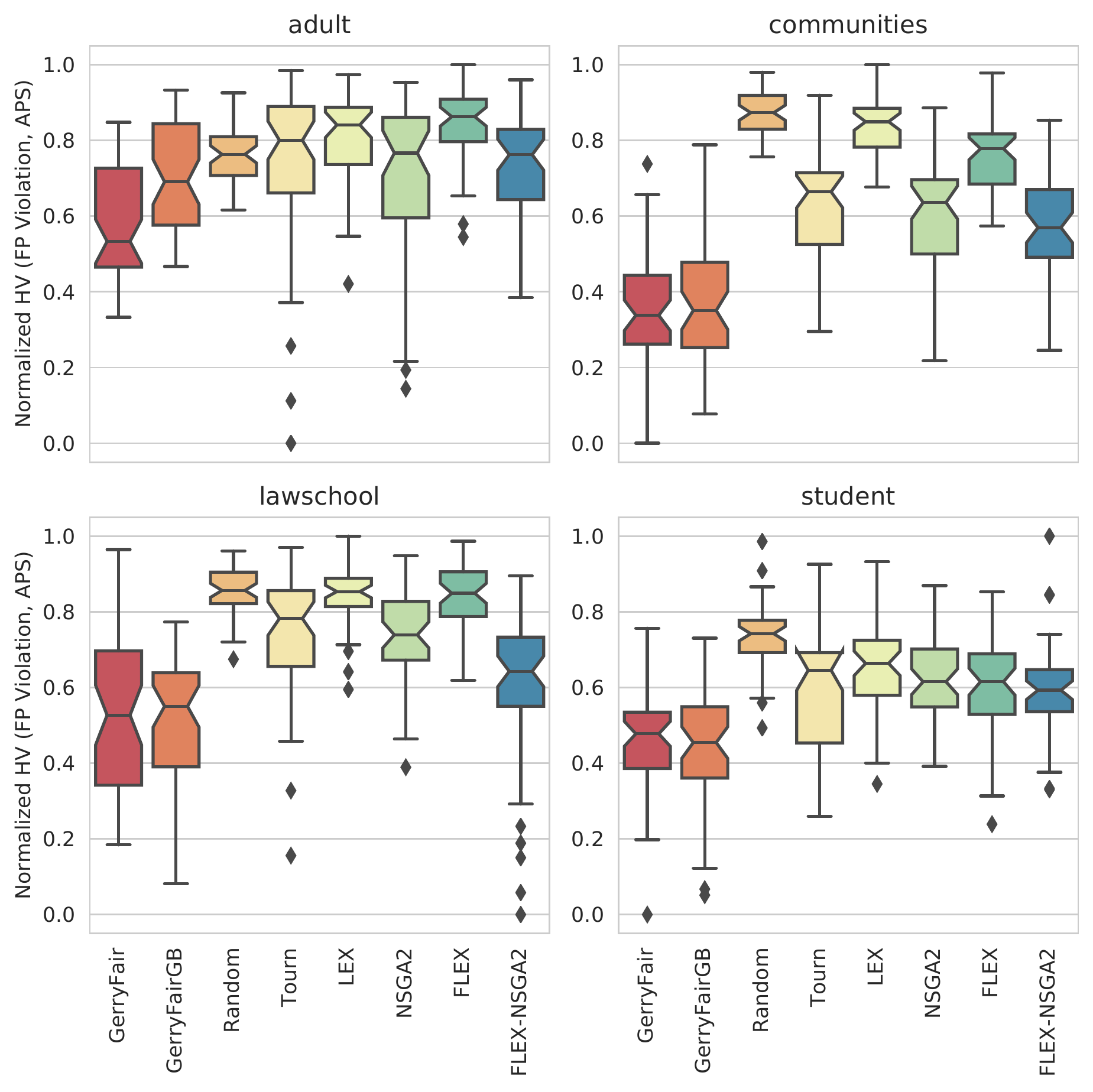}
    \caption{Normalized hypervolume of the Pareto front for test values of FP violation and average precision score.}
    \label{fig:fp_aps}
\end{figure}

In Figure~\ref{fig:fp_aps}, we show the distributions of the hypervolume of the FP violation-APS Pareto front across trials and problems for each method. 
Each subplot shows the test results for each method on a single dataset, with larger values indicating better performance. 
In general, we observe that the GP-based approaches do quite well compared to GerryFair in terms of finding good trade-offs along the Pareto front. 
Every GP variant generates a higher median hypervolume measure than GerryFair and GerryFairGB on every problem. 

Among GP variants, we observe that Random, LEX and FLEX tend to produce the highest hypervolume measures. 
Random search works best on the Communities and Student datasets; LEX performs best on Adult, and there is a virtual tie between Random, LEX and FLEX on Lawschool.
NSGA2, FLEX-NSGA2 and Tourn all perform similarly and generally worse than Random, LEX and FLEX. 

\begin{figure}
    \includegraphics[width=\columnwidth]{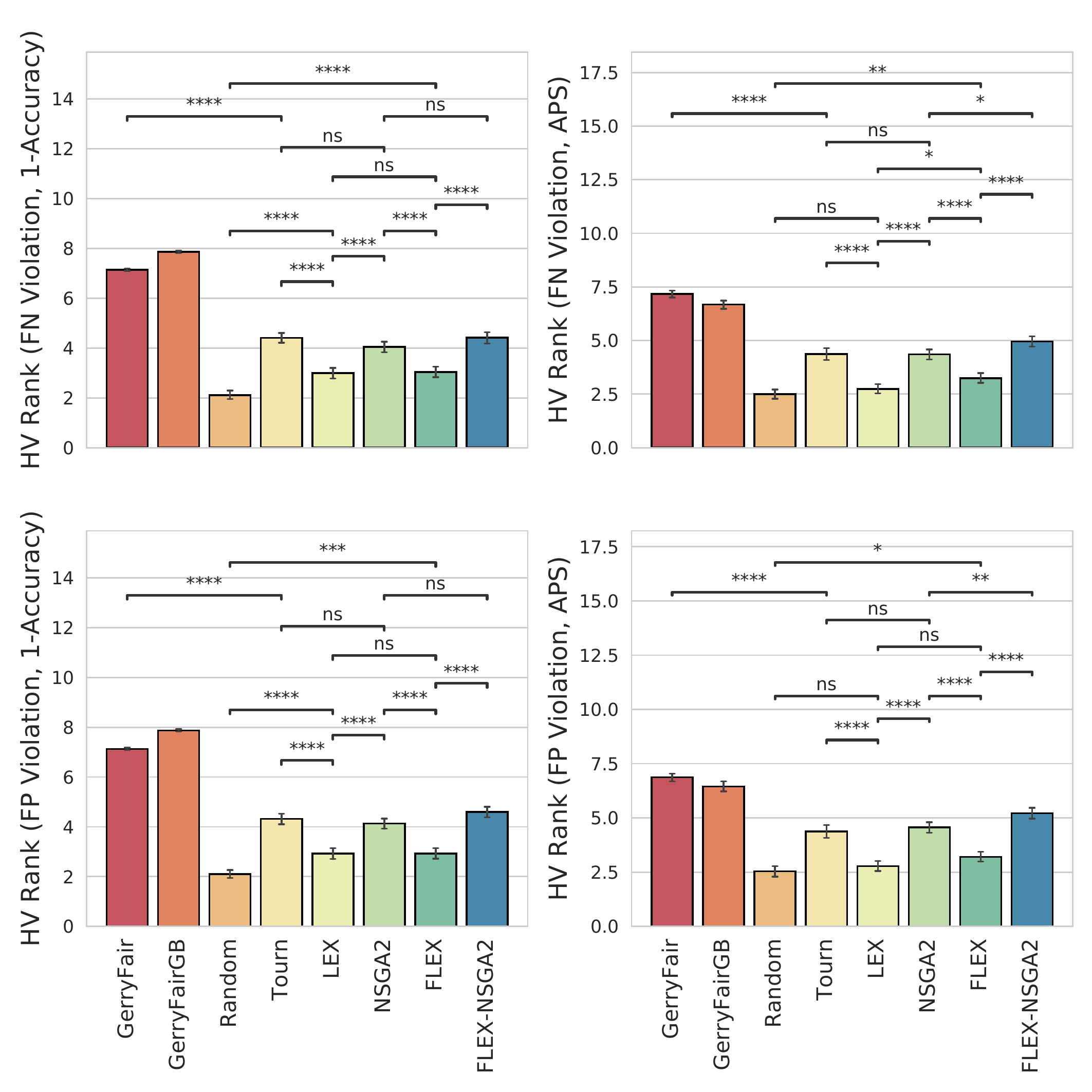}
    \caption{
        Rankings of methods by four different hypervolume (HV) measurements, across all problems.
        Asterisks denote statistical comparisons, conducted by a corrected pairwise Wilcoxon test. 
        ns: $5e-02 < p <= 1.0$;
        *: $1e-02 < p <= 5e-02$;
        **: $1e-03 < p <= 1e-02$;
        ***: $1e-04 < p <= 1e-03$;
        ****: $p <= 1e-04$.
    }
    \label{fig:ranks}
\end{figure}

The hypervolume performance results are further summarized across problems in Figure~\ref{fig:ranks}.
Here, each subplot shows the distribution of rankings according to a different hypervolume measurement, shown on the y axis. 
The significance of pairwise Wilcoxon tests between methods are shown as asterisks between bar plots. 
Since all pairwise comparisons are cumbersome to show, the complete pairwise Wilcoxon tests for FP Violation-APS hypervolume are shown in Table~\ref{tbl:stats}, corresponding to the bottom right subplot of Figure~\ref{fig:ranks}.

In general, the differences in performance between methods are significant. 
We observe that Random search, which has the best rankings across hypervolume measures, significantly outperforms all methods but LEX across problems. 
LEX and FLEX are significantly different only by one comparison, and the effect size is noticeably small. 
In addition, Tourn and NSGA2 are not significantly different, while NSGA2 and FLEX-NSGA2 are significantly different for two of the four measures. 

\begin{figure}
    \includegraphics[width=\columnwidth]{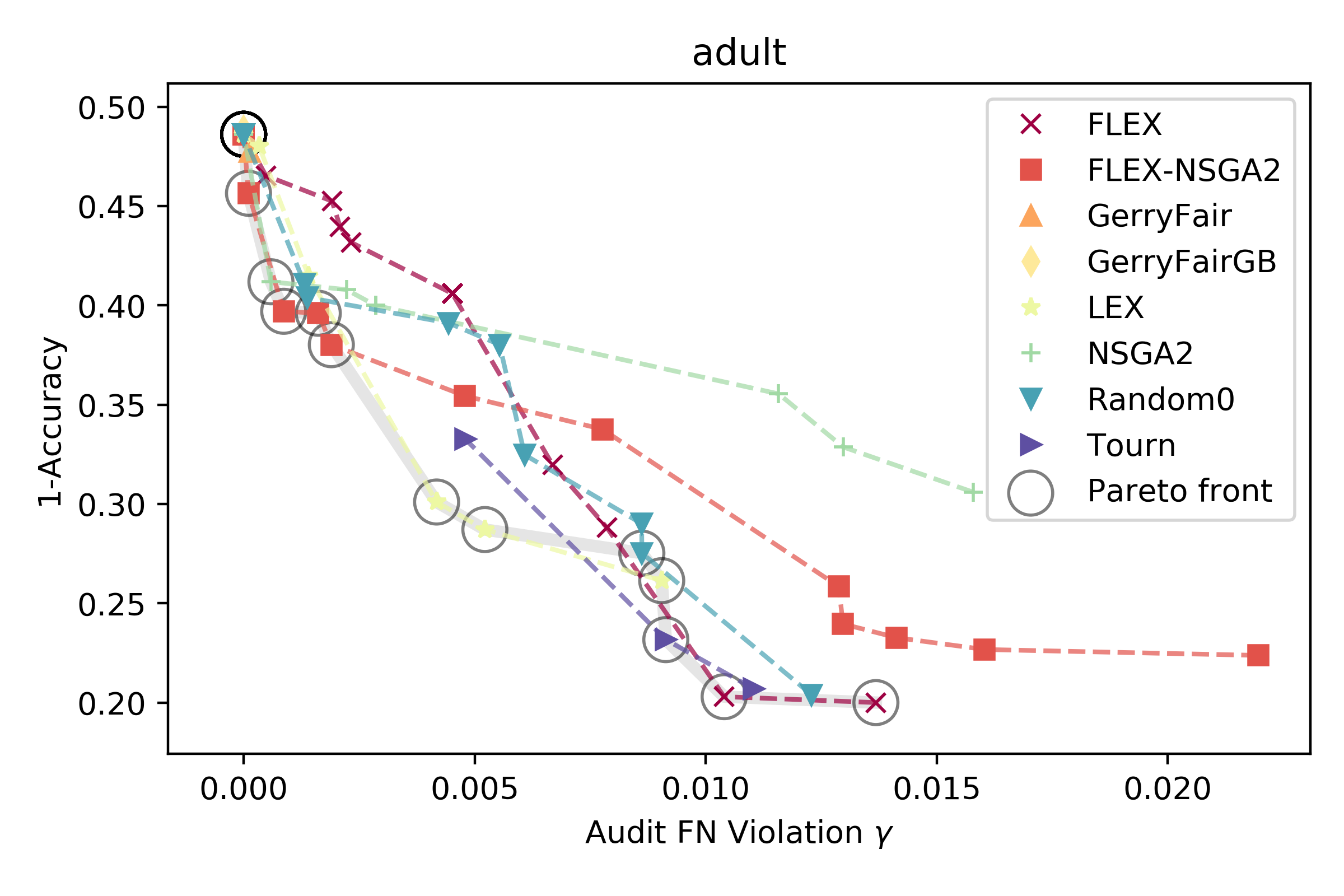}
    \caption{
        An example Pareto front of error (1-Accuracy) and unfairness (Audit FN Violation) based on test predictions on the adult dataset. 
        The test set Pareto fronts for each method are plotted separately with dotted lines.
        The combined Pareto front is circled, and consists of models from six different methods in this case. 
    }
    \label{fig:pf1}
\end{figure}

\begin{figure}
    \includegraphics[width=\columnwidth]{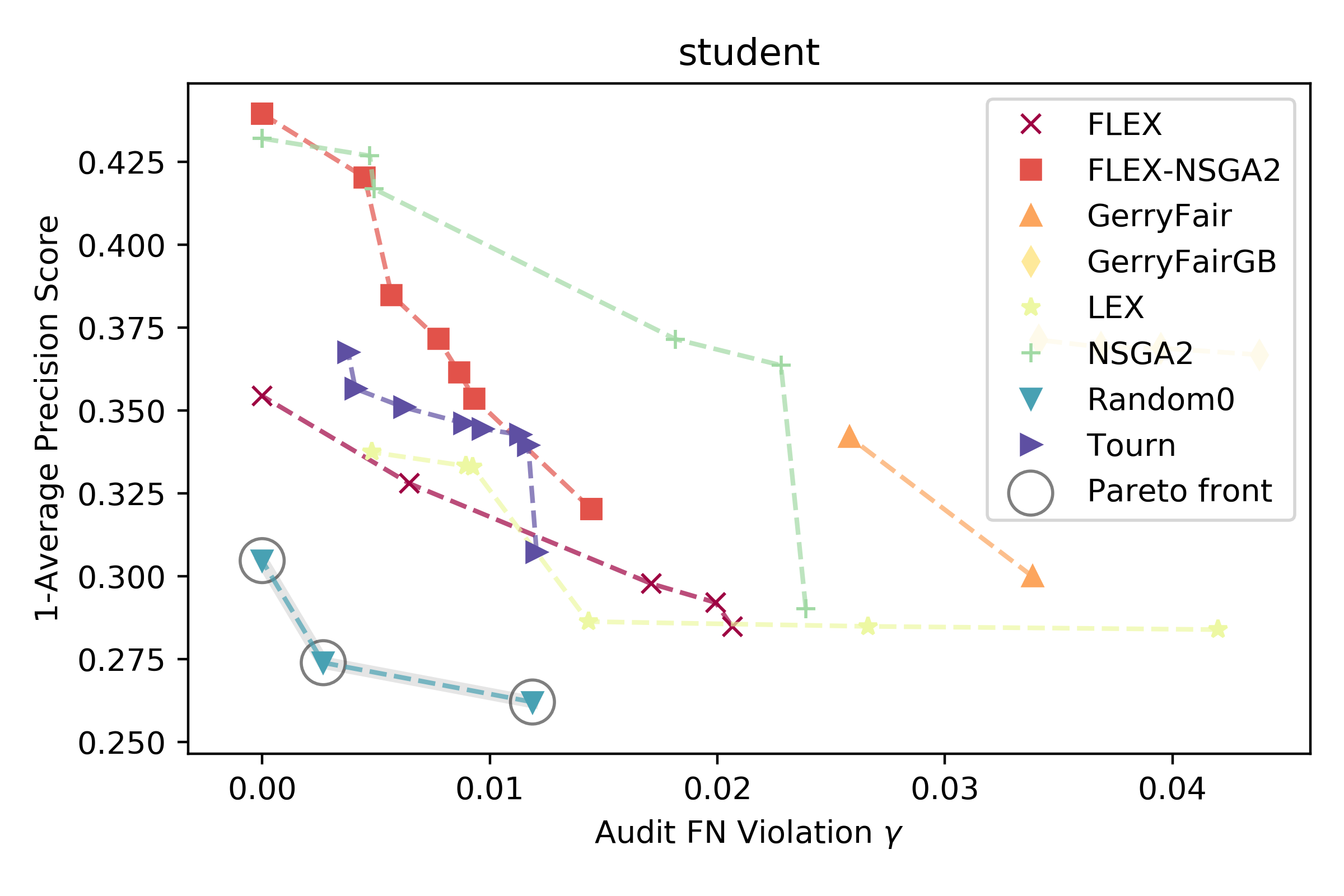}
    \caption{
        An example Pareto front of error (APS) and unfairness (Audit FN Violation) based on test predictions on the student dataset. 
        The test set Pareto fronts are plotted for each method separately with dotted lines.
        The combined Pareto front is circled, and consists of three models generated by random search that dominate all other models. 
    }
    \label{fig:pf2}
\end{figure}

Since the hypervolume measures only give a coarse grained view of what the Pareto fronts of solutions look like, we plot the Pareto fronts of specific trials of each method on two problems in Figures~\ref{fig:pf1} and~\ref{fig:pf2}. 
The first figure shows results for the Adult problem, and presents a typical solution set for this problem. 
It's noteworthy that, despite having 100 models produced by each method, only a fraction of these models produce Pareto-efficient sets on the test data. 
The small numbers of Pareto optimal models under test evaluation suggest that most classifiers are overfit to the training data to some degree, in terms of error rate, unfairness, or both.  
We also find it interesting that the combined front of solutions to this problem samples includes models from six different methods. 
In this way we see the potential for generating complimentary, Pareto-optimal models from distinct methods. 

By contrast, models for the Student dataset shown in Figure~\ref{fig:pf2} are dominated by one method: Random search. 
Random produces high hypervolume measures for this problem compared to other methods, and the Pareto fronts in this figure shows an example: in this case, Random is able to find three Pareto-optimal classifiers with very low error (high APS) and very low unfairness. 
These three models dominate all other solutions found by other methods.

\begin{figure}
    \includegraphics[width=0.8\columnwidth]{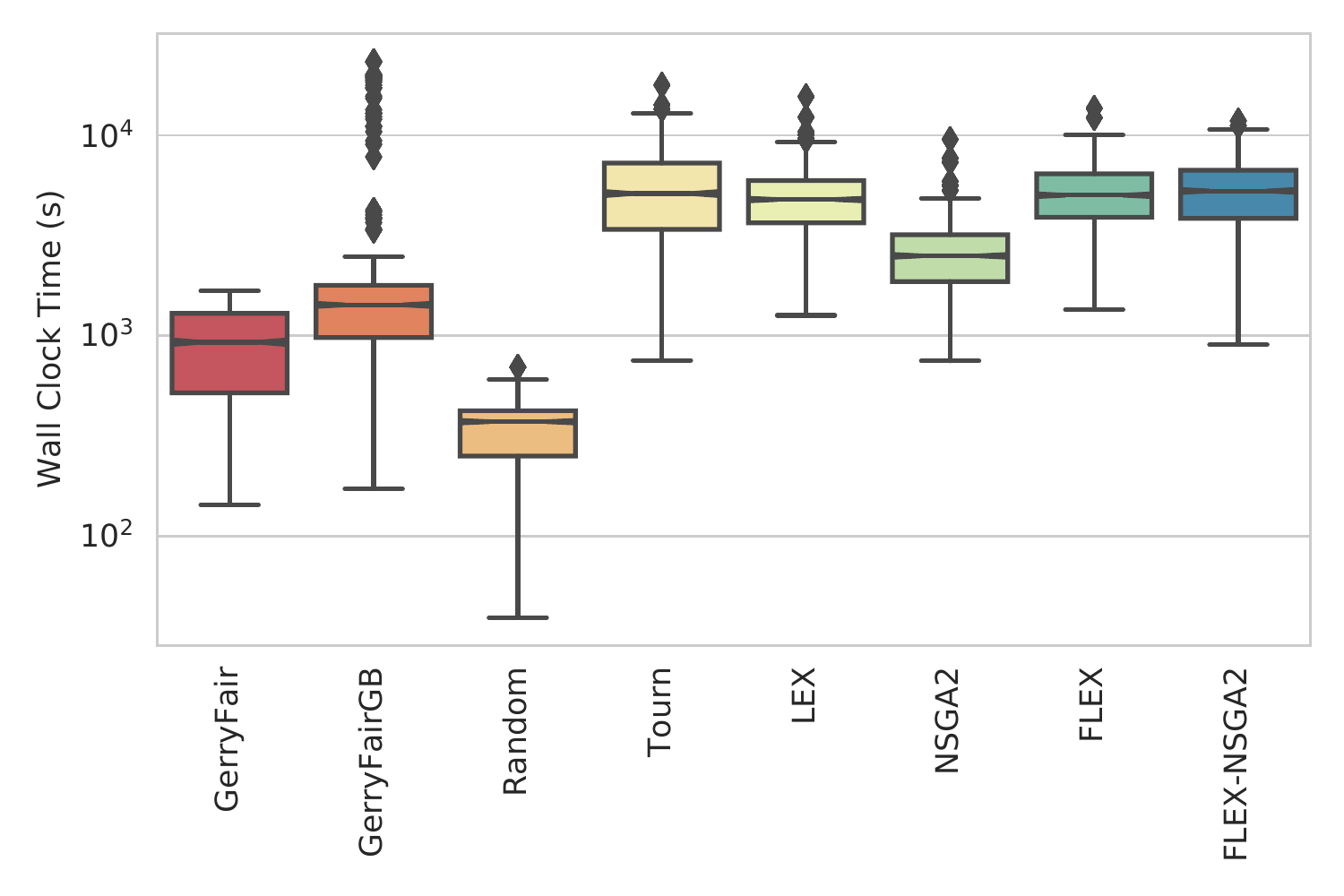}
    \caption{
        Wall clock runtime comparisons for all methods across all datasets.
    }
    \label{fig:time}
\end{figure}

Each method is evaluated on a single core, and the wall clock times of these runs are shown in Figure~\ref{fig:time}. 
Random is the quickest to train, followed by the two GerryFair variants.
Compared to the generational GP methods, GerryFair exhibits runtimes that are between 2 and 5 times faster.
Interestingly, the NSGA2 runs finish most quickly among the GP methods. 
This suggests that NSGA2 may be biased toward smaller models during optimization. 


\begin{table*}[htb]
\centering
\footnotesize
\caption{Bonferroni-adjusted $p$-values using a Wilcoxon signed rank test of (FP-Violation, APS) hypervolume scores for the methods across all problems. Bold: $p<$0.05.} 
\label{tbl:stats}
\begin{tabular}{rlllllll}
  \hline
 & FLEX & FLEX-NSGA2 & GerryFair & GerryFairGB & LEX & NSGA2 & Random \\ 
  \hline
FLEX-NSGA2 & {\bf 2.6e-16} &  &  &  &  &  &  \\ 
  GerryFair & {\bf 1.3e-30} & {\bf 7.4e-15} &  &  &  &  &  \\ 
  GerryFairGB & {\bf 4.2e-27} & {\bf 9.1e-10} & {\bf 3.1e-02} &  &  &  &  \\ 
  LEX & 1.5e-01 & {\bf 1.7e-20} & {\bf 7.9e-32} & {\bf 9.5e-27} &  &  &  \\ 
  NSGA2 & {\bf 1.7e-09} & {\bf 5.7e-03} & {\bf 5.6e-21} & {\bf 1.9e-14} & {\bf 2.1e-13} &  &  \\ 
  Random & {\bf 1.7e-02} & {\bf 1.3e-22} & {\bf 5.9e-32} & {\bf 3.5e-27} & 1.0e+00 & {\bf 3.7e-18} &  \\ 
  Tourn & {\bf 3.5e-06} & {\bf 2.3e-03} & {\bf 5.2e-20} & {\bf 7.3e-16} & {\bf 3.7e-11} & 1.0e+00 & {\bf 2.4e-12} \\ 
   \hline
\end{tabular}
\end{table*}


\section{Discussion and Conclusion}
\label{s:disuss}

The purpose of this work is to propose and evaluate methods for training fair classifiers using GP. 
We proposed two main ideas: first, to incorporate a fairness objective into NSGA2, and second, to modify lexicase selection to operate over subgroups of the protected attributes at each case, rather than on the raw samples. 
We evaluated these proposals relative to baseline GP approaches, including tournament selection, lexicase selection, and random search, and relative to a game-theoretic approach from the literature. 
In general we found that the GP-based methods perform quite well, in terms of the hypervolume dominated by the Pareto front of accuracy-fairness trade-offs they generate. 
An additional advantage of this family of methods is that they may generate intelligible models, due to their symbolic nature.
However, the typical evolutionary strategies used by GP did not perform significantly better than randomly generated models, except for one tested problem.

Our first idea, to incorporate a marginal fairness objective into NSGA2, did not result in model sets that were better than tournament selection. 
This suggests that the marginal fairness objective (Eq.~\ref{eq:marg_fair}) does not, in and of itself, produce model sets with better subgroup fairness (Eq.~\ref{eq:audit}). 
An obvious next step would be to incorporate the auditor (Section~\ref{s:auditor}) into NSGA2 in order to explicitly minimize the subgroup fitness violation.
The downside to this is its computational complexity, since it would require an additional iteration of model training per individual per generation. 

Our proposal to modify lexicase selection by regrouping cases in order to promote fairness over subgroups (FLEX) did not significantly change the performance of lexicase selection.
It appeared to improve performance on one dataset (adult), worsen performance on another (communities), and overall did not perform significantly differently than LEX.  
This comparison is overshadowed by the performance of random search over these datasets, which gave comparable, and occasionally better, performance than LEX and FLEX for a fraction of the computational cost. 

In light of these results, we want to understand why random search is so effective on these problems. 
There are several possible avenues of investigation. 
For one, the stability of the unfairness estimate provided by the auditor on training and test sets should be understood, since the group experiencing the largest fairness violation may differ between the two. 
This difference may make the fairness violation differ dramatically on test data.   
Unlike typical uses of Pareto optimization in GP literature that seek to control some static aspect of the solution (e.g., its complexity), in application to fairness, the risk of overfitting exists for both objectives. 
Therefore the robustness of Pareto optimal solutions may suffer.
In addition, the study conducted here considered a small number of small datasets, and it is possible that a larger number of datasets would reveal new and/or different insights.

The field of fairness in ML is nascent but growing quickly, and addresses very important societal concerns. 
Recent results show that the problems of both learning and auditing classifiers for rich subgroup fairness are computationally hard in the worst case. 
This motivates the analysis of heuristic algorithms such as GP for building fair classifiers. 
Our experiments suggest that GP-based methods can perform competitively with methods designed specifically for handling fairness. 
We hope that this motivates further inquiry into incorporating fairness constraints into models using randomized search heuristics such as evolutionary computation.

\section{Supplemental Material}
The code to reproduce our experiments is available from \url{https://github.com/lacava/fair_gp}.

\section{Acknowledgments}
The authors would like thank the Warren Center for Data Science and the Institute for Biomedical Informatics at Penn for their discussions. 
This work is supported by National Institutes of Health grants K99 LM012926-02, R01 LM010098 and R01 AI116794.

\bibliographystyle{ACM-Reference-Format}
\bibliography{Fairness} 


\begin{thebibliography}{29}


\ifx \showCODEN    \undefined \def \showCODEN     #1{\unskip}     \fi
\ifx \showDOI      \undefined \def \showDOI       #1{#1}\fi
\ifx \showISBNx    \undefined \def \showISBNx     #1{\unskip}     \fi
\ifx \showISBNxiii \undefined \def \showISBNxiii  #1{\unskip}     \fi
\ifx \showISSN     \undefined \def \showISSN      #1{\unskip}     \fi
\ifx \showLCCN     \undefined \def \showLCCN      #1{\unskip}     \fi
\ifx \shownote     \undefined \def \shownote      #1{#1}          \fi
\ifx \showarticletitle \undefined \def \showarticletitle #1{#1}   \fi
\ifx \showURL      \undefined \def \showURL       {\relax}        \fi
\providecommand\bibfield[2]{#2}
\providecommand\bibinfo[2]{#2}
\providecommand\natexlab[1]{#1}
\providecommand\showeprint[2][]{arXiv:#2}

\bibitem[\protect\citeauthoryear{Agarwal, Beygelzimer, Dudik, Langford, and
  Wallach}{Agarwal et~al\mbox{.}}{2018}]%
        {agarwal_reductions_2018}
\bibfield{author}{\bibinfo{person}{Alekh Agarwal}, \bibinfo{person}{Alina
  Beygelzimer}, \bibinfo{person}{Miroslav Dudik}, \bibinfo{person}{John
  Langford}, {and} \bibinfo{person}{Hanna Wallach}.}
  \bibinfo{year}{2018}\natexlab{}.
\newblock \showarticletitle{A {Reductions} {Approach} to {Fair}
  {Classification}}. In \bibinfo{booktitle}{{\em International {Conference} on
  {Machine} {Learning}}}. \bibinfo{pages}{60--69}.
\newblock
\showURL{%
\url{http://proceedings.mlr.press/v80/agarwal18a.html}}


\bibitem[\protect\citeauthoryear{Berk, Heidari, Jabbari, Joseph, Kearns,
  Morgenstern, Neel, and Roth}{Berk et~al\mbox{.}}{2017}]%
        {berk_convex_2017}
\bibfield{author}{\bibinfo{person}{Richard Berk}, \bibinfo{person}{Hoda
  Heidari}, \bibinfo{person}{Shahin Jabbari}, \bibinfo{person}{Matthew Joseph},
  \bibinfo{person}{Michael Kearns}, \bibinfo{person}{Jamie Morgenstern},
  \bibinfo{person}{Seth Neel}, {and} \bibinfo{person}{Aaron Roth}.}
  \bibinfo{year}{2017}\natexlab{}.
\newblock \showarticletitle{A convex framework for fair regression}.
\newblock \bibinfo{journal}{{\em arXiv preprint arXiv:1706.02409\/}}
  (\bibinfo{year}{2017}).
\newblock


\bibitem[\protect\citeauthoryear{Berk, Heidari, Jabbari, Kearns, and Roth}{Berk
  et~al\mbox{.}}{2018}]%
        {berk_fairness_2018}
\bibfield{author}{\bibinfo{person}{Richard Berk}, \bibinfo{person}{Hoda
  Heidari}, \bibinfo{person}{Shahin Jabbari}, \bibinfo{person}{Michael Kearns},
  {and} \bibinfo{person}{Aaron Roth}.} \bibinfo{year}{2018}\natexlab{}.
\newblock \showarticletitle{Fairness in {Criminal} {Justice} {Risk}
  {Assessments}: {The} {State} of the {Art}}.
\newblock \bibinfo{journal}{{\em Sociological Methods \& Research\/}}
  (\bibinfo{date}{July} \bibinfo{year}{2018}),
  \bibinfo{pages}{004912411878253}.
\newblock
\showISSN{0049-1241, 1552-8294}
\showDOI{%
\url{https://doi.org/10.1177/0049124118782533}}


\bibitem[\protect\citeauthoryear{Chand and Wagner}{Chand and Wagner}{2015}]%
        {chand_evolutionary_2015}
\bibfield{author}{\bibinfo{person}{Shelvin Chand} {and} \bibinfo{person}{Markus
  Wagner}.} \bibinfo{year}{2015}\natexlab{}.
\newblock \showarticletitle{Evolutionary many-objective optimization: {A}
  quick-start guide}.
\newblock \bibinfo{journal}{{\em Surveys in Operations Research and Management
  Science\/}} \bibinfo{volume}{20}, \bibinfo{number}{2} (\bibinfo{date}{Dec.}
  \bibinfo{year}{2015}), \bibinfo{pages}{35--42}.
\newblock
\showISSN{1876-7354}
\showDOI{%
\url{https://doi.org/10.1016/j.sorms.2015.08.001}}


\bibitem[\protect\citeauthoryear{Chouldechova and Roth}{Chouldechova and
  Roth}{2018}]%
        {chouldechova_frontiers_2018}
\bibfield{author}{\bibinfo{person}{Alexandra Chouldechova} {and}
  \bibinfo{person}{Aaron Roth}.} \bibinfo{year}{2018}\natexlab{}.
\newblock \showarticletitle{The {Frontiers} of {Fairness} in {Machine}
  {Learning}}.
\newblock \bibinfo{journal}{{\em arXiv:1810.08810 [cs, stat]\/}}
  (\bibinfo{date}{Oct.} \bibinfo{year}{2018}).
\newblock
\showURL{%
\url{http://arxiv.org/abs/1810.08810}}
\newblock
\shownote{arXiv: 1810.08810.}


\bibitem[\protect\citeauthoryear{Corbett-Davies, Pierson, Feller, Goel, and
  Huq}{Corbett-Davies et~al\mbox{.}}{2017}]%
        {corbett-davies_algorithmic_2017}
\bibfield{author}{\bibinfo{person}{Sam Corbett-Davies}, \bibinfo{person}{Emma
  Pierson}, \bibinfo{person}{Avi Feller}, \bibinfo{person}{Sharad Goel}, {and}
  \bibinfo{person}{Aziz Huq}.} \bibinfo{year}{2017}\natexlab{}.
\newblock \showarticletitle{Algorithmic decision making and the cost of
  fairness}. In \bibinfo{booktitle}{{\em Proceedings of the 23rd {ACM} {SIGKDD}
  {International} {Conference} on {Knowledge} {Discovery} and {Data}
  {Mining}}}. \bibinfo{publisher}{ACM}, \bibinfo{pages}{797--806}.
\newblock


\bibitem[\protect\citeauthoryear{Deb, Agrawal, Pratap, and Meyarivan}{Deb
  et~al\mbox{.}}{2000}]%
        {schoenauer_fast_2000}
\bibfield{author}{\bibinfo{person}{Kalyanmoy Deb}, \bibinfo{person}{Samir
  Agrawal}, \bibinfo{person}{Amrit Pratap}, {and} \bibinfo{person}{T
  Meyarivan}.} \bibinfo{year}{2000}\natexlab{}.
\newblock \showarticletitle{A {Fast} {Elitist} {Non}-dominated {Sorting}
  {Genetic} {Algorithm} for {Multi}-objective {Optimization}: {NSGA}-{II}}.
\newblock In \bibinfo{booktitle}{{\em Parallel {Problem} {Solving} from
  {Nature} {PPSN} {VI}}}, \bibfield{editor}{\bibinfo{person}{Marc Schoenauer},
  \bibinfo{person}{Kalyanmoy Deb}, \bibinfo{person}{G{\"u}nther Rudolph},
  \bibinfo{person}{Xin Yao}, \bibinfo{person}{Evelyne Lutton},
  \bibinfo{person}{Juan~Julian Merelo}, {and} \bibinfo{person}{Hans-Paul
  Schwefel}} (Eds.). Vol.~\bibinfo{volume}{1917}. \bibinfo{publisher}{Springer
  Berlin Heidelberg}, \bibinfo{address}{Berlin, Heidelberg},
  \bibinfo{pages}{849--858}.
\newblock
\showISBNx{978-3-540-41056-0}
\showURL{%
\url{http://repository.ias.ac.in/83498/}}


\bibitem[\protect\citeauthoryear{Dwork, Hardt, Pitassi, Reingold, and
  Zemel}{Dwork et~al\mbox{.}}{2012}]%
        {dwork_fairness_2012}
\bibfield{author}{\bibinfo{person}{Cynthia Dwork}, \bibinfo{person}{Moritz
  Hardt}, \bibinfo{person}{Toniann Pitassi}, \bibinfo{person}{Omer Reingold},
  {and} \bibinfo{person}{Richard Zemel}.} \bibinfo{year}{2012}\natexlab{}.
\newblock \showarticletitle{Fairness through awareness}. In
  \bibinfo{booktitle}{{\em Proceedings of the 3rd innovations in theoretical
  computer science conference}}. \bibinfo{publisher}{ACM},
  \bibinfo{pages}{214--226}.
\newblock


\bibitem[\protect\citeauthoryear{Fonseca, Paquete, and
  L{\'o}pez-Ib{\'a}nez}{Fonseca et~al\mbox{.}}{2006}]%
        {fonseca_improved_2006}
\bibfield{author}{\bibinfo{person}{Carlos~M. Fonseca},
  \bibinfo{person}{Lu{\'i}s Paquete}, {and} \bibinfo{person}{Manuel
  L{\'o}pez-Ib{\'a}nez}.} \bibinfo{year}{2006}\natexlab{}.
\newblock \showarticletitle{An improved dimension-sweep algorithm for the
  hypervolume indicator}. In \bibinfo{booktitle}{{\em 2006 {IEEE} international
  conference on evolutionary computation}}. \bibinfo{publisher}{IEEE},
  \bibinfo{pages}{1157--1163}.
\newblock


\bibitem[\protect\citeauthoryear{Gianfrancesco, Tamang, Yazdany, and
  Schmajuk}{Gianfrancesco et~al\mbox{.}}{2018}]%
        {gianfrancesco_potential_2018}
\bibfield{author}{\bibinfo{person}{Milena~A. Gianfrancesco},
  \bibinfo{person}{Suzanne Tamang}, \bibinfo{person}{Jinoos Yazdany}, {and}
  \bibinfo{person}{Gabriela Schmajuk}.} \bibinfo{year}{2018}\natexlab{}.
\newblock \showarticletitle{Potential biases in machine learning algorithms
  using electronic health record data}.
\newblock \bibinfo{journal}{{\em JAMA internal medicine\/}}
  \bibinfo{volume}{178}, \bibinfo{number}{11} (\bibinfo{year}{2018}),
  \bibinfo{pages}{1544--1547}.
\newblock


\bibitem[\protect\citeauthoryear{Hardt, Price, and Srebro}{Hardt
  et~al\mbox{.}}{2016}]%
        {hardt_equality_2016}
\bibfield{author}{\bibinfo{person}{Moritz Hardt}, \bibinfo{person}{Eric Price},
  {and} \bibinfo{person}{Nathan Srebro}.} \bibinfo{year}{2016}\natexlab{}.
\newblock \showarticletitle{Equality of {Opportunity} in {Supervised}
  {Learning}}.
\newblock  (\bibinfo{date}{Oct.} \bibinfo{year}{2016}).
\newblock
\showURL{%
\url{https://arxiv.org/abs/1610.02413v1}}


\bibitem[\protect\citeauthoryear{Helmuth, Pantridge, and Spector}{Helmuth
  et~al\mbox{.}}{2019}]%
        {helmuth_lexicase_2019}
\bibfield{author}{\bibinfo{person}{Thomas Helmuth}, \bibinfo{person}{Edward
  Pantridge}, {and} \bibinfo{person}{Lee Spector}.}
  \bibinfo{year}{2019}\natexlab{}.
\newblock \showarticletitle{Lexicase selection of specialists}. In
  \bibinfo{booktitle}{{\em Proceedings of the {Genetic} and {Evolutionary}
  {Computation} {Conference}}}. \bibinfo{pages}{1030--1038}.
\newblock


\bibitem[\protect\citeauthoryear{Helmuth, Spector, and Matheson}{Helmuth
  et~al\mbox{.}}{2014}]%
        {helmuth_solving_2014}
\bibfield{author}{\bibinfo{person}{T. Helmuth}, \bibinfo{person}{L. Spector},
  {and} \bibinfo{person}{J. Matheson}.} \bibinfo{year}{2014}\natexlab{}.
\newblock \showarticletitle{Solving {Uncompromising} {Problems} with {Lexicase}
  {Selection}}.
\newblock \bibinfo{journal}{{\em IEEE Transactions on Evolutionary
  Computation\/}} \bibinfo{volume}{PP}, \bibinfo{number}{99}
  (\bibinfo{year}{2014}), \bibinfo{pages}{1--1}.
\newblock
\showISSN{1089-778X}
\showDOI{%
\url{https://doi.org/10.1109/TEVC.2014.2362729}}


\bibitem[\protect\citeauthoryear{Kearns, Neel, Roth, and Wu}{Kearns
  et~al\mbox{.}}{2017}]%
        {kearns_preventing_2017}
\bibfield{author}{\bibinfo{person}{Michael Kearns}, \bibinfo{person}{Seth
  Neel}, \bibinfo{person}{Aaron Roth}, {and} \bibinfo{person}{Zhiwei~Steven
  Wu}.} \bibinfo{year}{2017}\natexlab{}.
\newblock \showarticletitle{Preventing {Fairness} {Gerrymandering}: {Auditing}
  and {Learning} for {Subgroup} {Fairness}}.
\newblock \bibinfo{journal}{{\em arXiv:1711.05144 [cs]\/}}
  (\bibinfo{date}{Nov.} \bibinfo{year}{2017}).
\newblock
\showURL{%
\url{http://arxiv.org/abs/1711.05144}}
\newblock
\shownote{arXiv: 1711.05144.}


\bibitem[\protect\citeauthoryear{Kearns, Neel, Roth, and Wu}{Kearns
  et~al\mbox{.}}{2018}]%
        {kearns_empirical_2018}
\bibfield{author}{\bibinfo{person}{Michael Kearns}, \bibinfo{person}{Seth
  Neel}, \bibinfo{person}{Aaron Roth}, {and} \bibinfo{person}{Zhiwei~Steven
  Wu}.} \bibinfo{year}{2018}\natexlab{}.
\newblock \showarticletitle{An {Empirical} {Study} of {Rich} {Subgroup}
  {Fairness} for {Machine} {Learning}}.
\newblock \bibinfo{journal}{{\em arXiv:1808.08166 [cs, stat]\/}}
  (\bibinfo{date}{Aug.} \bibinfo{year}{2018}).
\newblock
\showURL{%
\url{http://arxiv.org/abs/1808.08166}}
\newblock
\shownote{arXiv: 1808.08166.}


\bibitem[\protect\citeauthoryear{Kearns and Roth}{Kearns and Roth}{2019}]%
        {kearns_ethical_2019}
\bibfield{author}{\bibinfo{person}{Michael Kearns} {and} \bibinfo{person}{Aaron
  Roth}.} \bibinfo{year}{2019}\natexlab{}.
\newblock \bibinfo{booktitle}{{\em The {Ethical} {Algorithm}: {The} {Science}
  of {Socially} {Aware} {Algorithm} {Design}}}.
\newblock \bibinfo{publisher}{Oxford University Press}.
\newblock


\bibitem[\protect\citeauthoryear{Kleinberg, Mullainathan, and
  Raghavan}{Kleinberg et~al\mbox{.}}{2016}]%
        {kleinberg_inherent_2016}
\bibfield{author}{\bibinfo{person}{Jon Kleinberg}, \bibinfo{person}{Sendhil
  Mullainathan}, {and} \bibinfo{person}{Manish Raghavan}.}
  \bibinfo{year}{2016}\natexlab{}.
\newblock \showarticletitle{Inherent trade-offs in the fair determination of
  risk scores}.
\newblock \bibinfo{journal}{{\em arXiv preprint arXiv:1609.05807\/}}
  (\bibinfo{year}{2016}).
\newblock


\bibitem[\protect\citeauthoryear{La~Cava, Helmuth, Spector, and Moore}{La~Cava
  et~al\mbox{.}}{2018}]%
        {la_cava_probabilistic_2018}
\bibfield{author}{\bibinfo{person}{William La~Cava}, \bibinfo{person}{Thomas
  Helmuth}, \bibinfo{person}{Lee Spector}, {and} \bibinfo{person}{Jason~H.
  Moore}.} \bibinfo{year}{2018}\natexlab{}.
\newblock \showarticletitle{A probabilistic and multi-objective analysis of
  lexicase selection and $\varepsilon$-lexicase selection}.
\newblock \bibinfo{journal}{{\em Evolutionary Computation\/}}
  (\bibinfo{date}{May} \bibinfo{year}{2018}), \bibinfo{pages}{1--28}.
\newblock
\showISSN{1063-6560}
\showDOI{%
\url{https://doi.org/10.1162/evco_a_00224}}


\bibitem[\protect\citeauthoryear{La~Cava and Moore}{La~Cava and Moore}{2019}]%
        {la_cava_semantic_2019}
\bibfield{author}{\bibinfo{person}{William La~Cava} {and}
  \bibinfo{person}{Jason~H. Moore}.} \bibinfo{year}{2019}\natexlab{}.
\newblock \showarticletitle{Semantic variation operators for multidimensional
  genetic programming}. In \bibinfo{booktitle}{{\em Proceedings of the 2019
  {Genetic} and {Evolutionary} {Computation} {Conference}}} {\em
  (\bibinfo{series}{{GECCO} '19})}. \bibinfo{publisher}{ACM},
  \bibinfo{address}{Prague, Czech Republic}.
\newblock
\showDOI{%
\url{https://doi.org/10.1145/3321707.3321776}}
\newblock
\shownote{arXiv: 1904.08577.}


\bibitem[\protect\citeauthoryear{La~Cava and Moore}{La~Cava and Moore}{2020}]%
        {lacava_learning_2020}
\bibfield{author}{\bibinfo{person}{William La~Cava} {and}
  \bibinfo{person}{Jason~H. Moore}.} \bibinfo{year}{2020}\natexlab{}.
\newblock \showarticletitle{Learning feature spaces for regression with genetic
  programming}.
\newblock \bibinfo{journal}{{\em Genetic Programming and Evolvable Machines\/}}
  (\bibinfo{date}{March} \bibinfo{year}{2020}).
\newblock
\showISSN{1573-7632}
\showDOI{%
\url{https://doi.org/10.1007/s10710-020-09383-4}}


\bibitem[\protect\citeauthoryear{La~Cava, Singh, Taggart, Suri, and
  Moore}{La~Cava et~al\mbox{.}}{2019}]%
        {la_cava_learning_2019}
\bibfield{author}{\bibinfo{person}{William La~Cava}, \bibinfo{person}{Tilak~Raj
  Singh}, \bibinfo{person}{James Taggart}, \bibinfo{person}{Srinivas Suri},
  {and} \bibinfo{person}{Jason~H. Moore}.} \bibinfo{year}{2019}\natexlab{}.
\newblock \showarticletitle{Learning concise representations for regression by
  evolving networks of trees}. In \bibinfo{booktitle}{{\em International
  {Conference} on {Learning} {Representations}}} {\em
  (\bibinfo{series}{{ICLR}})}.
\newblock
\showURL{%
\url{https://arxiv.org/abs/1807.00981}}


\bibitem[\protect\citeauthoryear{La~Cava, Spector, and Danai}{La~Cava
  et~al\mbox{.}}{2016}]%
        {la_cava_epsilon-lexicase_2016}
\bibfield{author}{\bibinfo{person}{William La~Cava}, \bibinfo{person}{Lee
  Spector}, {and} \bibinfo{person}{Kourosh Danai}.}
  \bibinfo{year}{2016}\natexlab{}.
\newblock \showarticletitle{Epsilon-{Lexicase} {Selection} for {Regression}}.
  In \bibinfo{booktitle}{{\em Proceedings of the {Genetic} and {Evolutionary}
  {Computation} {Conference} 2016}} {\em (\bibinfo{series}{{GECCO} '16})}.
  \bibinfo{publisher}{ACM}, \bibinfo{address}{New York, NY, USA},
  \bibinfo{pages}{741--748}.
\newblock
\showISBNx{978-1-4503-4206-3}
\showDOI{%
\url{https://doi.org/10.1145/2908812.2908898}}


\bibitem[\protect\citeauthoryear{Liskowski, Krawiec, Helmuth, and
  Spector}{Liskowski et~al\mbox{.}}{2015}]%
        {liskowski_comparison_2015}
\bibfield{author}{\bibinfo{person}{Pawel Liskowski}, \bibinfo{person}{Krzysztof
  Krawiec}, \bibinfo{person}{Thomas Helmuth}, {and} \bibinfo{person}{Lee
  Spector}.} \bibinfo{year}{2015}\natexlab{}.
\newblock \showarticletitle{Comparison of {Semantic}-aware {Selection}
  {Methods} in {Genetic} {Programming}}. In \bibinfo{booktitle}{{\em
  Proceedings of the {Companion} {Publication} of the 2015 {Annual}
  {Conference} on {Genetic} and {Evolutionary} {Computation}}} {\em
  (\bibinfo{series}{{GECCO} {Companion} '15})}. \bibinfo{publisher}{ACM},
  \bibinfo{address}{New York, NY, USA}, \bibinfo{pages}{1301--1307}.
\newblock
\showISBNx{978-1-4503-3488-4}
\showDOI{%
\url{https://doi.org/10.1145/2739482.2768505}}


\bibitem[\protect\citeauthoryear{Marcinkowski, Kieslich, Starke, and
  L{\"u}nich}{Marcinkowski et~al\mbox{.}}{2020}]%
        {marcinkowski_implications_2020}
\bibfield{author}{\bibinfo{person}{Frank Marcinkowski}, \bibinfo{person}{Kimon
  Kieslich}, \bibinfo{person}{Christopher Starke}, {and} \bibinfo{person}{Marco
  L{\"u}nich}.} \bibinfo{year}{2020}\natexlab{}.
\newblock \showarticletitle{Implications of {AI} (un-)fairness in higher
  education admissions: the effects of perceived {AI} (un-)fairness on exit,
  voice and organizational reputation}. In \bibinfo{booktitle}{{\em Proceedings
  of the 2020 {Conference} on {Fairness}, {Accountability}, and
  {Transparency}}} {\em (\bibinfo{series}{{FAT}* '20})}.
  \bibinfo{publisher}{Association for Computing Machinery},
  \bibinfo{address}{Barcelona, Spain}, \bibinfo{pages}{122--130}.
\newblock
\showISBNx{978-1-4503-6936-7}
\showDOI{%
\url{https://doi.org/10.1145/3351095.3372867}}


\bibitem[\protect\citeauthoryear{Pedreshi, Ruggieri, and Turini}{Pedreshi
  et~al\mbox{.}}{2008}]%
        {pedreshi_discrimination-aware_2008}
\bibfield{author}{\bibinfo{person}{Dino Pedreshi}, \bibinfo{person}{Salvatore
  Ruggieri}, {and} \bibinfo{person}{Franco Turini}.}
  \bibinfo{year}{2008}\natexlab{}.
\newblock \showarticletitle{Discrimination-aware data mining}. In
  \bibinfo{booktitle}{{\em Proceedings of the 14th {ACM} {SIGKDD} international
  conference on {Knowledge} discovery and data mining}}.
  \bibinfo{pages}{560--568}.
\newblock


\bibitem[\protect\citeauthoryear{Quadrianto and Sharmanska}{Quadrianto and
  Sharmanska}{2017}]%
        {quadrianto_recycling_2017}
\bibfield{author}{\bibinfo{person}{Novi Quadrianto} {and}
  \bibinfo{person}{Viktoriia Sharmanska}.} \bibinfo{year}{2017}\natexlab{}.
\newblock \showarticletitle{Recycling privileged learning and distribution
  matching for fairness}. In \bibinfo{booktitle}{{\em Advances in {Neural}
  {Information} {Processing} {Systems}}}. \bibinfo{pages}{677--688}.
\newblock


\bibitem[\protect\citeauthoryear{Spector}{Spector}{2012}]%
        {spector_assessment_2012}
\bibfield{author}{\bibinfo{person}{Lee Spector}.}
  \bibinfo{year}{2012}\natexlab{}.
\newblock \showarticletitle{Assessment of problem modality by differential
  performance of lexicase selection in genetic programming: a preliminary
  report}. In \bibinfo{booktitle}{{\em Proceedings of the fourteenth
  international conference on {Genetic} and evolutionary computation conference
  companion}}. \bibinfo{pages}{401--408}.
\newblock
\showURL{%
\url{http://dl.acm.org/citation.cfm?id=2330846}}


\bibitem[\protect\citeauthoryear{Thomas, Silva, Barto, Giguere, Brun, and
  Brunskill}{Thomas et~al\mbox{.}}{2019}]%
        {thomas_preventing_2019}
\bibfield{author}{\bibinfo{person}{Philip~S. Thomas}, \bibinfo{person}{Bruno
  Castro~da Silva}, \bibinfo{person}{Andrew~G. Barto}, \bibinfo{person}{Stephen
  Giguere}, \bibinfo{person}{Yuriy Brun}, {and} \bibinfo{person}{Emma
  Brunskill}.} \bibinfo{year}{2019}\natexlab{}.
\newblock \showarticletitle{Preventing undesirable behavior of intelligent
  machines}.
\newblock \bibinfo{journal}{{\em Science\/}} \bibinfo{volume}{366},
  \bibinfo{number}{6468} (\bibinfo{date}{Nov.} \bibinfo{year}{2019}),
  \bibinfo{pages}{999--1004}.
\newblock
\showISSN{0036-8075, 1095-9203}
\showDOI{%
\url{https://doi.org/10.1126/science.aag3311}}


\bibitem[\protect\citeauthoryear{Zink and Rose}{Zink and Rose}{2019}]%
        {zink_fair_2019}
\bibfield{author}{\bibinfo{person}{Anna Zink} {and} \bibinfo{person}{Sherri
  Rose}.} \bibinfo{year}{2019}\natexlab{}.
\newblock \showarticletitle{Fair {Regression} for {Health} {Care} {Spending}}.
\newblock \bibinfo{journal}{{\em arXiv:1901.10566 [cs, stat]\/}}
  (\bibinfo{date}{Jan.} \bibinfo{year}{2019}).
\newblock
\showURL{%
\url{http://arxiv.org/abs/1901.10566}}
\newblock
\shownote{arXiv: 1901.10566.}


\end{thebibliography}

\end{document}